\documentclass[conference]{IEEEtran}
\IEEEoverridecommandlockouts
\usepackage{cite}
\usepackage{amsmath,amssymb,amsfonts}
\usepackage{graphicx}
\usepackage{textcomp}
\usepackage{xcolor}
\usepackage{multirow}
\usepackage{multicol}
\usepackage{graphicx}
\usepackage{booktabs}
\usepackage{algorithm}
\usepackage{colortbl}
\usepackage{algpseudocode}

\def\BibTeX{{\rm B\kern-.05em{\sc i\kern-.025em b}\kern-.08em
    T\kern-.1667em\lower.7ex\hbox{E}\kern-.125emX}}
\begin{document}

\title{Extending Straight-Through Estimation for Robust Neural Networks on Analog CIM Hardware}
\author{
    \IEEEauthorblockN{Yuannuo Feng\textsuperscript{\dag}$^{1,3}$, Wenyong Zhou\textsuperscript{\dag}$^{2,3}$, Yuexi Lyu\textsuperscript{\ddag}$^{3}$, Yixiang Zhang$^{3}$, Zhengwu Liu*$^{2}$, Ngai Wong*$^{2}$ and Wang Kang*$^{1}$}
    \IEEEauthorblockA{
        $^1$School of Integrated Circuit Science and Engineering, Beihang University, Beijing, China\\
        $^2$Department of Electrical and Electronic Engineering, The University of Hong Kong, Hong Kong\\
        $^3$Zhicun Research Lab, Beijing, China\\
       \textsuperscript{\dag}: Equal Contribution. \textsuperscript{\ddag}: Project Leader. *: Corresponding Author(s).
    }
}
\maketitle
\begin{abstract}
Analog Compute-In-Memory (CIM) architectures promise significant energy efficiency gains for neural network inference, but suffer from complex hardware-induced noise that poses major challenges for deployment. While noise-aware training methods have been proposed to address this issue, they typically rely on idealized and differentiable noise models that fail to capture the full complexity of analog CIM hardware variations. Motivated by the Straight-Through Estimator (STE) framework in quantization, we decouple forward noise simulation from backward gradient computation, enabling noise-aware training with more accurate but computationally intractable noise modeling in analog CIM systems. We provide theoretical analysis demonstrating that our approach preserves essential gradient directional information while maintaining computational tractability and optimization stability. 
Extensive experiments show that our extended STE framework achieves up to 5.3\% accuracy improvement on image classification, 0.72 perplexity reduction on text generation, 2.2$\times$ speedup in training time, and 37.9\% lower peak memory usage compared to standard noise-aware training methods.
\end{abstract}

\section{Introduction}
\label{sec:introduction}
The exponential growth of neural network applications has intensified demand for energy-efficient computing solutions, particularly for edge devices with severe power and computational constraints~\cite{icsict, asicon}. Analog Compute-In-Memory (CIM) architectures address these challenges by performing matrix-vector multiplications directly within memory arrays, eliminating energy-intensive data movement and achieving orders of magnitude energy efficiency improvements over traditional von Neumann architectures through analog weight storage and physical law-based computation~\cite{shafiee2016isaac, prime}.

However, the analog nature introduces significant deployment challenges, as analog CIM hardware suffers from various noise sources that severely degrade inference accuracy~\cite{mao2025hyimc, pan2022mini}. These include device-level variations from manufacturing and aging, circuit-level noise from voltage drops and current leakage, and peripheral circuit imperfections such as ADC quantization errors and amplifier non-linearities~\cite{accurate, bayesft}. The cumulative effect of these noise sources can lead to substantial accuracy degradation, limiting practical neural network deployment on analog CIM hardware~\cite{chen2024device, bai2024end}.

To address this challenge, researchers have developed noise-aware training methods that expose neural networks to simulated hardware noise during training, enabling them to develop robustness against deployment conditions~\cite{zhoudate, icassp}. However, existing approaches face a fundamental limitation: they rely on simplified, differentiable noise models that can be easily integrated into standard gradient-based optimization frameworks. While Gaussian noise injection or uniform quantization can be readily handled by automatic differentiation, they fail to capture the full complexity of real analog CIM hardware variations, as many critical hardware effects are inherently non-differentiable, exhibit complex spatial and temporal correlations, or involve computationally expensive physical simulations that make direct gradient computation impractical~\cite{accurate}. This limitation creates a significant gap between the simplified noise models used during training and the complex noise characteristics encountered during deployment, where networks trained with oversimplified noise models often fail to generalize to real hardware conditions, resulting in substantial accuracy drops when deployed on actual analog CIM systems~\cite{batchnorm}.

We extend the Straight-Through Estimator (STE) framework, originally for quantization, to address complex noise in analog CIM hardware~\cite{nu2uq, birealnet}. By decoupling forward noise simulation from backward gradient computation, we enable realistic noise models in forward passes and simplified gradients in backpropagation, achieving efficient noise-aware training with high-fidelity hardware models. This paper provides three key contributions:
\begin{itemize}
    \item We formalize the extension of STE to complex noise environments, providing a general framework that can accommodate various types of hardware noise beyond simple quantization.
    \item We develop a STE-based gradient approximation strategy and provide theoretical understanding of its effectiveness from the gradient perspective.
    \item Experiments demonstrate that our extended STE framework achieves up to 5.3\% higher accuracy on image classification, 0.72 lower perplexity on text generation, 2.2$\times$ faster training, and 37.9\% less peak memory usage than standard noise-aware methods.
\end{itemize}

\section{Preliminaries}
\label{sec:preliminaries}
The STE was originally developed to enable gradient-based training of neural networks containing non-differentiable functions, particularly quantization operations~\cite{dsq}. In the standard quantization scenario, the forward pass applies a quantization function $Q(\cdot)$ to the weights or activations:
\begin{equation}
y = Q(x)
\end{equation}
where $Q(\cdot)$ is typically a step function that maps continuous values to discrete levels.

The challenge arises during backpropagation, as the gradient of $Q(\cdot)$ is zero almost everywhere, preventing effective learning. The STE addresses this by using different functions for forward and backward passes. During the forward pass, the actual quantization function is applied, while during the backward pass, the gradient flows through as if the quantization operation were the identity function:
\begin{equation}
\frac{\partial L}{\partial x} = \frac{\partial L}{\partial y} \cdot \mathbf{1}
\end{equation}
where $\mathbf{1}$ represents the identity operation~\cite{irnet}.

\section{Methodology}
\subsection{Problem Formulation}
We consider training neural networks that operate robustly under complex deployment noise. Let $\mathbf{W} \in \mathbb{R}^{d}$ denote the clean weights and $\mathcal{N}(\mathbf{W}, \boldsymbol{\xi})$ represent a complex noise function, where $\boldsymbol{\xi}$ encapsulates noise parameters. The noisy weights are:
\begin{equation}
\tilde{\mathbf{W}} = \mathcal{N}(\mathbf{W}, \boldsymbol{\xi})
\end{equation}

The training objective minimizes the expected loss:
\begin{equation}
\mathbb{E}_{\boldsymbol{\xi}}[L] = \mathbb{E}_{\boldsymbol{\xi}}[\ell(f(\tilde{\mathbf{W}}, \mathbf{x}), \mathbf{y})]
\end{equation}

The challenge arises when computing gradients $\frac{\partial L}{\partial \mathbf{W}}$ through $\mathcal{N}(\mathbf{W}, \boldsymbol{\xi})$, which may be non-differentiable, have zero gradients almost everywhere, or involve computationally expensive stochastic processes that make standard backpropagation infeasible.

Algorithm~\ref{alg:noisy_linear} demonstrates our STE approach for a linear layer. During the forward pass, inputs and weights are quantized to int8 precision, converted to appropriate formats for the noise model, and processed to generate noisy outputs. The difference between the noisy and clean outputs is computed as a delta term, which is then added to the clean output using gradient detachment to preserve backpropagation through the original linear computation while incorporating the effects of quantization noise.
\subsection{Theoretical Analysis}
To understand how the STE can be extended to complex noise environments, we analyze the approach from the perspective of gradient direction and magnitude. The effectiveness of STE can be understood by decomposing any gradient vector $\mathbf{g}$ into its components:
\begin{equation}
\mathbf{g} = \|\mathbf{g}\| \cdot \frac{\mathbf{g}}{\|\mathbf{g}\|}
\end{equation}
where $\|\mathbf{g}\|$ represents the gradient norm (step size information) and $\frac{\mathbf{g}}{\|\mathbf{g}\|}$ represents the gradient direction (update direction).
\begin{algorithm}[!t]
\caption{Noisy Linear Layer with Quantization and Noise}
\label{alg:noisy_linear}
\begin{algorithmic}[1]
\Function{Linear}{$x$, $linear$, $noise\_level$, $with\_noise$}
    \If{not $with\_noise$}
        \State \Return $\text{Linear}(x, linear.weight, linear.bias)$
    \EndIf
    
    \State $y_{clean} = \text{Linear}(x, linear.weight, linear.bias)$
    \State $noise\_model = \text{NoiseModelNNTrain}(noise\_level)$
    
    \State \textbf{with} $\text{torch.no\_grad()}$:
        \State \quad $x_{q8}, scale_x = \text{Quant}_{int8}(x)$
        \State \quad $w_{q8}, scale_w = \text{Quant}_{int8}(linear.weight)$
        \State \quad $x_{uint8}, w_{split} = \text{SplitInput}(x_{q8}, w_{q8})$
        \State \quad $bias_{q} = \text{QuantBias}(linear.bias, scale_x, scale_w)$
        \State \quad $g = \text{ComputeGain}(scale_x, scale_w, y_{clean})$
        
        \State \quad $noise\_model.\text{program}(x_{uint8}, w_{split}, bias_{q}, g)$
        \State \quad $y_{noisy} = noise\_model.\text{infer}(x_{uint8})$
        \State \quad $y_{scaled} = \text{Rescale}(y_{noisy}, g, scale_x, scale_w)$
        \State \quad $\delta = y_{scaled} - y_{clean}.\text{detach}()$
    
    \State \Return $y_{clean} + \delta.\text{detach}()$
\EndFunction
\end{algorithmic}
\end{algorithm}

Consider a neural network layer computation affected by complex hardware noise. The forward computation can be expressed as:
\begin{equation}
\mathbf{y} = \tilde{\mathbf{W}} \mathbf{x} = \mathcal{N}(\mathbf{W}, \boldsymbol{\xi}) \mathbf{x}
\end{equation}
where $\mathcal{N}(\cdot, \cdot)$ represents the complex noise function that transforms the clean weights $\mathbf{W}$ into noisy weights $\tilde{\mathbf{W}}$.

The ground truth gradient that properly accounts for the noise characteristics would be:
\begin{equation}
\mathbf{g}^* = \frac{\partial L}{\partial \mathbf{W}} = \frac{\partial L}{\partial \mathbf{y}} \frac{\partial \mathcal{N}(\mathbf{W}, \boldsymbol{\xi})}{\partial \mathbf{W}} \mathbf{x}
\end{equation}

However, computing this gradient exactly is often intractable due to the complexity and non-differentiable nature of $\mathcal{N}(\cdot, \cdot)$. The STE approach approximates this ground truth gradient using a simplified gradient that assumes $\frac{\partial \mathcal{N}(\mathbf{W}, \boldsymbol{\xi})}{\partial \mathbf{W}} \approx \mathbf{I}$:
\begin{equation}
\tilde{\mathbf{g}} = \frac{\partial L}{\partial \mathbf{y}} \mathbf{x}
\end{equation}

From a statistical perspective, we can view the relationship between the ground truth gradient and the STE gradient as:
\begin{equation}
\mathbf{g}^* = \tilde{\mathbf{g}} + \boldsymbol{\delta}
\end{equation}
where $\boldsymbol{\delta} = \frac{\partial L}{\partial \mathbf{y}} \left(\frac{\partial \mathcal{N}(\mathbf{W}, \boldsymbol{\xi})}{\partial \mathbf{W}} - \mathbf{I}\right) \mathbf{x}$ represents the bias term introduced by the noise function.

\textbf{Direction Analysis:} The STE gradient $\tilde{\mathbf{g}}$ represents the full-precision gradient direction, which captures the correct optimization trajectory for the clean network. To quantify the directional reliability, consider the cosine similarity between the STE gradient and the ground truth gradient:
\begin{equation}
\cos(\theta) = \frac{\langle \tilde{\mathbf{g}}, \mathbf{g}^* \rangle}{\|\tilde{\mathbf{g}}\| \|\mathbf{g}^*\|} = \frac{\langle \tilde{\mathbf{g}}, \tilde{\mathbf{g}} + \boldsymbol{\delta} \rangle}{\|\tilde{\mathbf{g}}\| \|\tilde{\mathbf{g}} + \boldsymbol{\delta}\|}
\end{equation}

Expanding this expression:
\begin{equation}
\cos(\theta) = \frac{\|\tilde{\mathbf{g}}\|^2 + \langle \tilde{\mathbf{g}}, \boldsymbol{\delta} \rangle}{\|\tilde{\mathbf{g}}\| \|\tilde{\mathbf{g}} + \boldsymbol{\delta}\|}
\end{equation}

When the noise-induced bias $\boldsymbol{\delta}$ is uncorrelated with the clean gradient $\tilde{\mathbf{g}}$ (i.e., $\mathbb{E}[\langle \tilde{\mathbf{g}}, \boldsymbol{\delta} \rangle] \approx 0$), the cosine similarity approaches:
\begin{equation}
\mathbb{E}[\cos(\theta)] \approx \frac{\|\tilde{\mathbf{g}}\|}{\|\tilde{\mathbf{g}} + \boldsymbol{\delta}\|} \approx \frac{\|\tilde{\mathbf{g}}\|}{\sqrt{\|\tilde{\mathbf{g}}\|^2 + \|\boldsymbol{\delta}\|^2}}
\end{equation}

This shows that the directional alignment deteriorates as the noise bias magnitude $\|\boldsymbol{\delta}\|$ increases relative to the clean gradient magnitude $\|\tilde{\mathbf{g}}\|$. The variance of the ground truth gradient can be expressed as:
\begin{equation}
\text{Var}[\mathbf{g}^*] = \text{Var}[\boldsymbol{\delta}] = \mathbb{E}[\|\boldsymbol{\delta}\|^2] - \|\mathbb{E}[\boldsymbol{\delta}]\|^2
\end{equation}

Since $\tilde{\mathbf{g}}$ is deterministic given the clean weights, it exhibits zero variance, making it statistically more reliable for consistent optimization direction.

The key insight is that $\tilde{\mathbf{g}}$ provides a projection of the optimization direction onto the clean parameter space. Mathematically, this can be viewed as:
\begin{equation}
\tilde{\mathbf{g}} = \mathbb{E}_{\boldsymbol{\xi}}[\mathbf{g}^*] \quad \text{when} \quad \mathbb{E}_{\boldsymbol{\xi}}[\boldsymbol{\delta}] = \mathbf{0}
\end{equation}

When noise is not zero-mean, $\tilde{\mathbf{g}}$ provides a simplified approximation of the optimization direction by ignoring the systematic bias introduced by the noise in $\mathbf{g}^*$.

\textbf{Magnitude Analysis:} While the magnitude of $\tilde{\mathbf{g}}$ may differ from $\mathbf{g}^*$, this discrepancy is less critical for optimization success. The magnitude relationship can be expressed as:
\begin{equation}
\|\mathbf{g}^*\|^2 = \|\tilde{\mathbf{g}} + \boldsymbol{\delta}\|^2 = \|\tilde{\mathbf{g}}\|^2 + 2\langle \tilde{\mathbf{g}}, \boldsymbol{\delta} \rangle + \|\boldsymbol{\delta}\|^2
\end{equation}

The relative magnitude difference is:
\begin{equation}
\frac{\|\mathbf{g}^*\|^2 - \|\tilde{\mathbf{g}}\|^2}{\|\tilde{\mathbf{g}}\|^2} = \frac{2\langle \tilde{\mathbf{g}}, \boldsymbol{\delta} \rangle + \|\boldsymbol{\delta}\|^2}{\|\tilde{\mathbf{g}}\|^2}
\end{equation}

Modern optimizers such as Adam effectively normalize gradients using adaptive scaling:
\begin{equation}
\mathbf{W}_{t+1} = \mathbf{W}_t - \alpha \frac{\hat{\mathbf{m}}_t}{\sqrt{\hat{\mathbf{v}}_t} + \epsilon}
\end{equation}
where $\hat{\mathbf{m}}_t$ and $\hat{\mathbf{v}}_t$ are bias-corrected momentum estimates. This adaptive mechanism makes the optimization relatively insensitive to the exact gradient magnitude, as the effective step size is determined by the gradient history rather than instantaneous magnitude.

\section{Experiments}
\label{sec:experiments}
\subsection{Experiment Setup}
We develop a comprehensive noise simulator modeling realistic analog CIM accelerator conditions with both I/O and tile-level non-idealities detailed in Table~\ref{tab:nonidealities}. I/O non-idealities include ADC/DAC quantization noise and device non-linearity effects. Tile-level non-idealities encompass programming noise from weight fabrication variations, cycle-by-cycle read variance, thermal-induced parameter drift, retention-based memory degradation, and IR-drop effects from wire resistance causing voltage variations across crossbar arrays.

\subsection{Experiment Result}
\begin{table}[!t]
\centering
\renewcommand{\arraystretch}{1.3}
\caption{Hardware non-idealities modeled in CIM noise simulator.}
\resizebox{\columnwidth}{!}{%
\begin{tabular}{lll}
\toprule
Category & Noise Source & Type \\
\midrule
\multirow{3}{*}{I/O non-idealities} & ADC noise & Quantization noise \\
& DAC noise & Quantization noise \\
& Device non-linearity & System non-linearity \\
\midrule
\multirow{5}{*}{Tile non-idealities} & Programming noise & Weight fabrication non-ideality \\
& Cycle-by-cycle read variance & Computational consistency \\
& Thermal variation & Temperature-dependent drift \\
& Retention & Memory degradation \\
& IR-drop & Wire resistance non-ideality \\
\bottomrule
\end{tabular}%
}
\label{tab:nonidealities}
\end{table}
\begin{table}[!t]
\renewcommand{\arraystretch}{1.5}
\caption{Performance Comparison of Models Across Different Lognormal Noise Levels.}
\centering
\resizebox{\columnwidth}{!}{
\begin{tabular}{llcccccc}
\toprule
\multirow{3}{*}{Dataset}      & \multirow{3}{*}{Model} & \multicolumn{6}{c}{Performance under Noise} \\ \cmidrule(lr){3-8}
                            &                         & \multicolumn{2}{c}{$level = 1.0$} & \multicolumn{2}{c}{$level = 2.0$} & \multicolumn{2}{c}{$level = 3.0$} \\ \cmidrule(lr){3-4} \cmidrule(lr){5-6} \cmidrule(lr){7-8}
                            &                         & Value & $\Delta$ & Value & $\Delta$ & Value & $\Delta$ \\ \hline
\multirow{4}{*}{\textbf{CIFAR-10}}      & VGG-8      & 90.14\% & \cellcolor{blue!10}{+1.43\%} & 89.56\% & \cellcolor{blue!10}{+1.99\%} & 89.38\% & \cellcolor{blue!10}{+2.75\%}\\
                            & VGG-11      & 89.90\% & \cellcolor{blue!10}{+3.20\%} & 89.56\% & \cellcolor{blue!10}{+4.50\%} & 89.90\% & \cellcolor{blue!10}{+5.30\%}\\
                            & ResNet-20      & 90.77\% & \cellcolor{blue!10}{+2.57\%} & 90.66\% & \cellcolor{blue!10}{+2.65\%} & 90.44\% & \cellcolor{blue!10}{+3.02\%}\\
                            & ResNet-18      & 94.00\% & \cellcolor{blue!10}{+3.58\%} & 93.59\% & \cellcolor{blue!10}{+1.98\%} & 93.78\% & \cellcolor{blue!10}{+2.56\%}\\ \cmidrule(lr){1-8}
\multirow{2}{*}{\textbf{Shakespeare}}      & BERT      & 3.10 & \cellcolor{red!20}{-0.31} & 3.11 & \cellcolor{red!20}{-0.33} & 3.16 & \cellcolor{red!20}{-0.24}\\
                            & GPT      & 4.65 & \cellcolor{red!20}{-0.51} & 4.69 & \cellcolor{red!20}{-0.50} & 4.69 & \cellcolor{red!20}{-0.72}\\
                            \bottomrule
\end{tabular}
}
\label{tab:result}
\end{table}
\begin{table}[!t]
\renewcommand{\arraystretch}{1.5}
\caption{Training Phase Hardware Cost Comparison of Different Gradient Management Methods.}
\centering
\resizebox{\columnwidth}{!}{
\begin{tabular}{lcccccc}
\toprule
\multirow{2}{*}{Method} & \multirow{2}{*}{\begin{tabular}{c}Time\\(s)\end{tabular}} & \multicolumn{2}{c}{Memory (MB)} & \multicolumn{2}{c}{GPU Utilization (\%)} \\ \cmidrule(lr){3-4} \cmidrule(lr){5-6}
                        &  & Peak & Average & Compute & Memory \\ \hline
Baseline                & 0.136 & 1501 & 342 & 64.9 & 4.14 \\
Full Gradient           & 1.172 & 4019 & 343 & 96.9 & 7.94 \\
\rowcolor{blue!10}
\texttt{torch.no\_grad} & 0.608  & 2496 & 342  & 94.5 & 8.44 \\
\texttt{.detach()}      & 0.616 & 3124 & 342  & 94.3 & 8.40 \\
\bottomrule
\end{tabular}
}
\label{tab:hardware_cost}
\end{table}

Table~\ref{tab:result} presents the performance comparison of various models across different lognormal noise levels, demonstrating that our method significantly improves performance under different noise conditions compared to baseline models. For CIFAR-10 classification tasks, all models show substantial accuracy increases, with improvements ranging from 1.43\% to 5.3\%. VGG-11 exhibits the most remarkable gains, achieving up to 5.3\% accuracy increase at noise level 3.0. For the Shakespeare language modeling task, both BERT and GPT models experience slight perplexity decreases, with reductions ranging from 0.24 to 0.72.

Table~\ref{tab:hardware_cost} presents a comprehensive comparison of training phase hardware costs across different gradient management methods. The baseline method, which represents standard training without complex noise considerations, demonstrates the most efficient resource utilization with the lowest time consumption (0.136s), minimal memory usage (1501MB peak, 342MB average). In contrast, the full gradient approach exhibits significantly higher computational overhead, requiring 1.172s per iteration and consuming substantially more memory (4019MB peak, 343MB average). The gradient management techniques \texttt{torch.no\_grad} and \texttt{.detach()} show similar performance characteristics, with execution times of 0.608s and 0.616s respectively, and comparable memory consumption patterns (2496MB and 3124MB peak usage, both with 342MB average). Our proposed methods effectively save memory and accelerate training speed compared to the full gradient approach, with \texttt{torch.no\_grad} emerging as the superior solution due to its optimal balance between computational efficiency and resource utilization.
\begin{figure}[!t]
\centering
\includegraphics[width=1.0\linewidth]{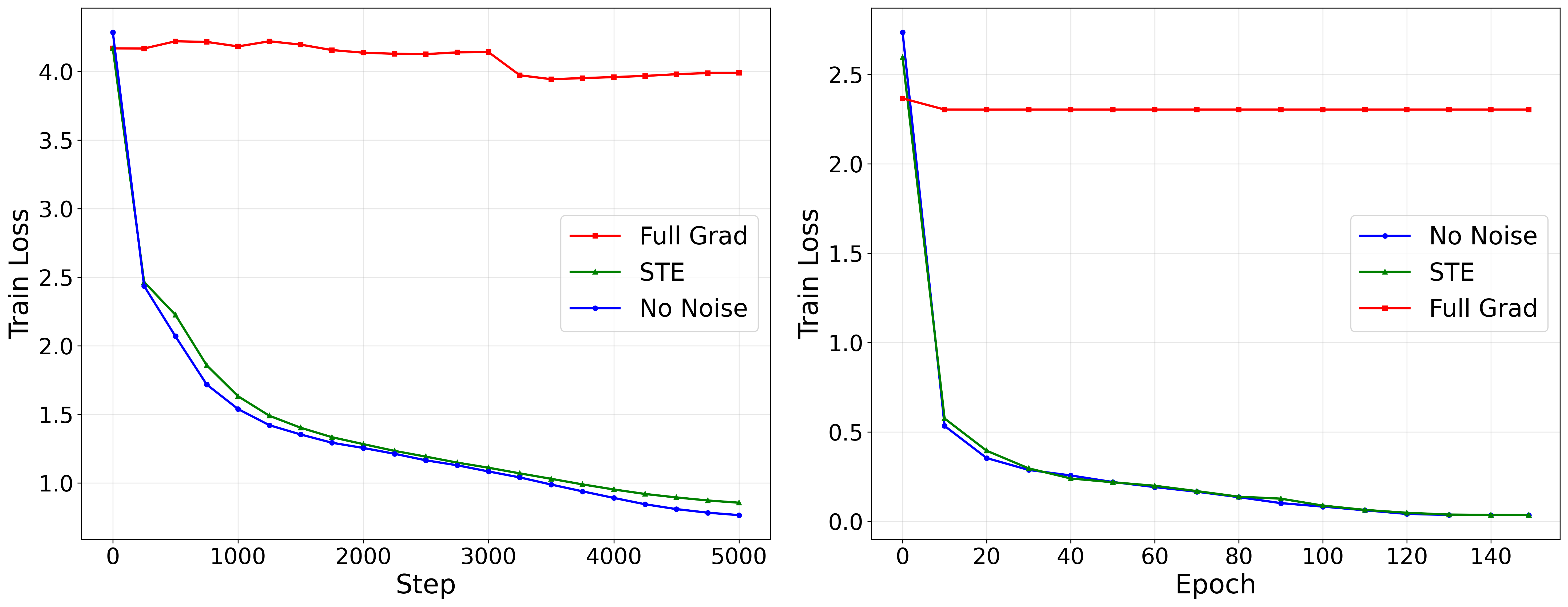}
\caption{The training loss reduction curves for GPT-2 on the Shakespeare dataset and ResNet on the CIFAR-10 dataset.}
\label{fig:training_validation_loss}
\end{figure}

The left panel of Fig.~\ref{fig:training_validation_loss} shows the training loss curves for GPT-2 models trained on the Shakespeare dataset using three different methods: baseline (No Noise), noise with gradients (Full Grad), and STE. Both the baseline and our STE method exhibit rapid convergence during the initial 1000 training steps, dropping from approximately 4.0 to below 2.0 and reaching final losses around 0.8-1.0. In contrast, the noise gradient method shows minimal improvement throughout training, maintaining consistently high loss around 4.0 for both training and validation sets, suggesting that directly propagating gradients through noisy operations significantly impairs the optimization process.
Similarly, the right panel shows ResNet’s performance on CIFAR-10 for image classification, highlighting a significant advantage of our STE method over full gradient training.
\section{Conclusion}
\label{sec:conclusion}
This work addresses the gap between realistic analog CIM noise modeling and practical training by extending the STE framework to handle complex, non-differentiable hardware variations. By decoupling forward noise simulation from backward gradient computation, our approach enables high-fidelity noise modeling while maintaining computational tractability. Our work improves image classification accuracy by 5.3\%, cuts text generation perplexity by 0.72, speeds training 2.2$\times$, and reduces peak memory use by 37.9\% versus standard noise-aware methods.
\section*{Acknowledgement}
\label{sec:acknowledgement}
This research was partially conducted by ACCESS – AI Chip Center for Emerging Smart Systems, supported by the InnoHK initiative of the Innovation and Technology Commission of the Hong Kong Special Administrative Region Government, and partially supported by the Theme-based Research Scheme (TRS) project T45-701/22-R, the General Research Fund (GRF) Project 17203224 of the Research Grants Council (RGC), Hong Kong SAR, and the National Natural Science Foundation of China Project 62404187.
\bibliographystyle{ieeetr}
\bibliography{bibliography}
\end{document}